 \documentclass[pmlr,twocolumn,10pt]{jmlr} % W&CP article

\usepackage{booktabs}
\usepackage{siunitx}

\usepackage[switch]{lineno}

\theorembodyfont{\upshape}
\theoremheaderfont{\scshape}
\theorempostheader{:}
\theoremsep{\newline}

\jmlrvolume{}
\jmlryear{2024}
\jmlrsubmitted{}
\jmlrpublished{}
\jmlrworkshop{Attribute Structuring for LLM-Based Evaluation} % W&CP title

\usepackage{graphicx}
\usepackage{float}

\usepackage{placeins} %FloatBarrier
\usepackage{multirow}
\usepackage[capitalize]{cleveref}
\usepackage{array}
\newcolumntype{P}[1]{>{\centering\arraybackslash}p{#1}}

\usepackage{cleveref}

\usepackage{fontawesome}

\usepackage{adjustbox}

\crefname{section}{Sec.}{Secs.}%
\usepackage{scrextend} % for addmargin
\usepackage{siunitx}

\usepackage{makecell}
\newcommand{\capitalizefirst}[1]{\MakeUppercase{\expandafter\@car#1\@nil}\expandafter\@cdr#1\@nil}

\usepackage{changepage}   % for the adjustwidth environment

\usepackage[ampersand]{easylist}
\ListProperties(Hide=100, Hang=true, Progressive=2ex, Style*=$\bullet$ , Style2*=$\circ$)

\usepackage{xspace}
\newcommand{\method}{Attribute Structuring\xspace}
\newcommand{\methodshort}{AS\xspace}

\title[Attribute Structuring Improves LLM-Based Evaluation of Clinical Text Summaries]{Attribute Structuring Improves LLM-Based Evaluation\\of Clinical Text Summaries}

 \author{%
\Name{Zelalem Gero}$^1$\Email{zelalemgero@microsoft.com}\\
\Name{Chandan Singh} $^1$\Email{chansingh@microsoft.com}\\
\Name{Yiqing Xie} $^2$\Email{yiqingxi@andrew.cmu.edu}\\
\Name{Sheng Zhang} $^1$\Email{zhang.sheng@microsoft.com}\\
\Name{Praveen Subramanian} $^3$\Email{praveen.s@microsoft.com}\\
 \Name{Paul Vozila} $^3$\Email{paulvozila@microsoft.com}\\
\Name{Tristan Naumann} $^1$\Email{tristan@microsoft.com}\\
\Name{Jianfeng Gao} $^1$\Email{jfgao@microsoft.com}\\
 \Name{Hoifung Poon} $^1$\Email{hoifung@microsoft.com}\\
\\{\text{\hspace{170pt}\centering \normalfont $^1$ Microsoft Research}}
\\{\text{\hspace{170pt}\centering \normalfont $^2$ Carnegie Mellon University}}
\\{\text{\hspace{170pt}\centering \normalfont $^3$ Microsoft Health and Life Sciences}}
 }

\begin{document}

\maketitle

\begin{abstract}
Summarizing clinical text is crucial in health decision-support and clinical research. Large language models (LLMs) have shown the potential to generate accurate clinical text summaries, but still struggle with issues regarding grounding and evaluation, especially in safety-critical domains such as health. Holistically evaluating text summaries is challenging because they may contain unsubstantiated information. Here, we explore a general mitigation framework using Attribute Structuring (AS), which structures the summary evaluation process. It decomposes the evaluation process into a grounded procedure that uses an LLM for relatively simple structuring and scoring tasks, rather than the full task of holistic summary evaluation. Experiments show that AS consistently improves the correspondence between human annotations and automated metrics in clinical text summarization. Additionally, AS yields interpretations in the form of a short text span corresponding to each output, which enables efficient human auditing, paving the way towards trustworthy evaluation of clinical information in resource-constrained scenarios. 
\end{abstract}


\paragraph*{Data and Code Availability}
We release our full code at \url{https://github.com/microsoft/attribute-structuring}.
All analyzed data comes from the freely available discharge notes in \mbox{MIMIC-III}~\citep{johnson2016mimic}.

% \paragraph*{Institutional Review Board (IRB)}
% IRB information will be provided upon acceptance.

\section{Introduction}

Automatic medical text summarization has emerged as a critical task for real-world impact in natural-language processing amidst widespread physician fatigue and burnout.
Meanwhile, pre-trained large language models (LLMs) have shown impressive proficiency in a range of complex natural language processing tasks~\citep{brown2020language,dubey2024llama,openai2023gpt4}, including in text summarization~\citep{liu2022revisiting, zhang2024benchmarking}.
However, evaluating these summaries is challenging, as they rely on human effort to track different attributes and facts.

% To address this issue, we propose a method that uses LLMs to evaluating medical text summaries, guided by a simple structuring process.
LLMs may be able to help automate this evaluation.
LLMs have previously been used to provide practical evaluations in a variety of evaluation settings~\citep{liu2023gpteval, fu2023gptscore}.
However, these evaluations have been imperfect: yielding relatively low correlation with human judgments,
showing bias~\citep{zheng2023judging,wang2023automated},
or more generally struggling with long-form nuanced evaluation, as is often the case in summarization~\citep{zhong2022towards}.

\begin{figure*}
    \centering
    \includegraphics[width=0.8\textwidth]{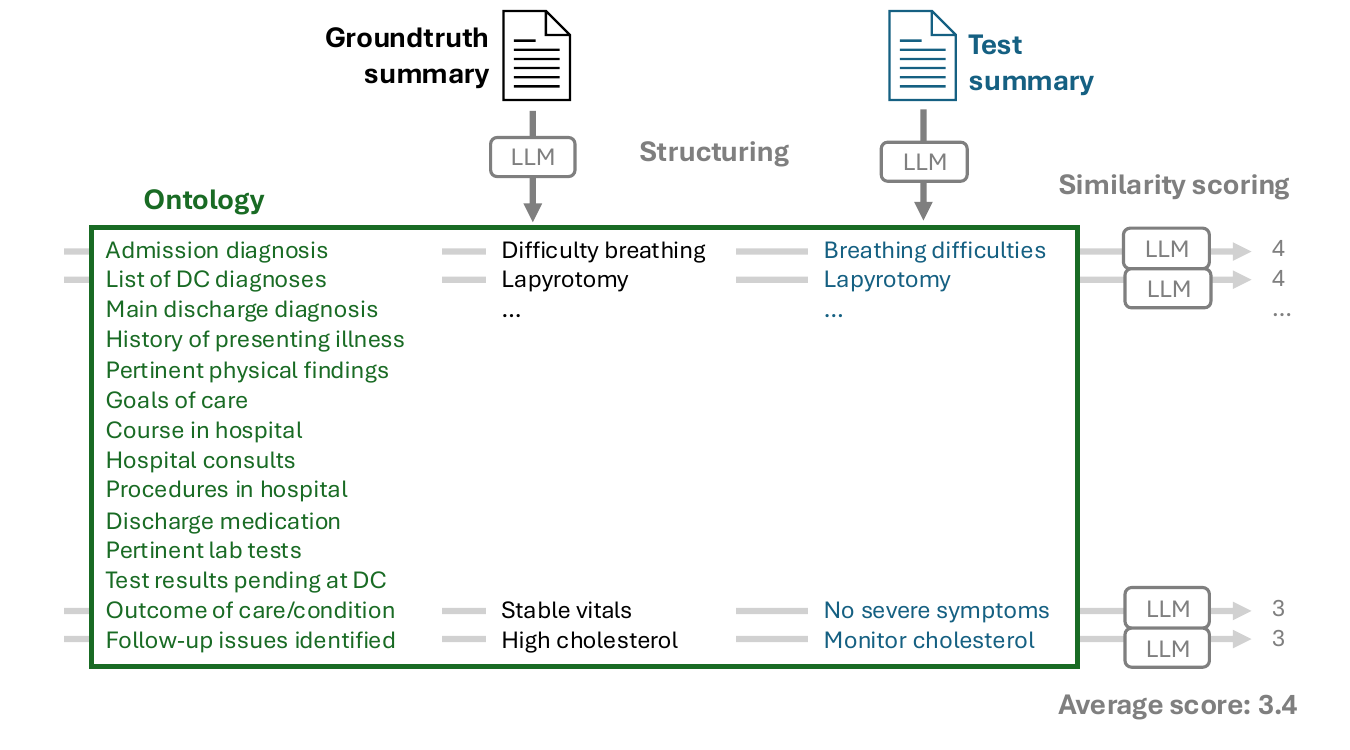}
    \vspace{-10pt}
    \caption{
    Overview of \method LLM evaluation for clinical summarization.
    \methodshort uses a human-given ontology to extract attributes from summaries and then to individually score them.
    }
    \label{fig:intro}
\end{figure*}

% \looseness=-1
To address these issues, structuring the scoring process can help to ground the evaluation, and help an LLM decompose a complex evaluation process into simpler, more accurate steps.
Following this intuition, we introduce a simple structuring process, which we call \method{} (\methodshort) (\cref{fig:intro}).
Given a groundtruth summary and a summary to be evaluated, \methodshort leverages an LLM to score a summary through prompting.
Rather than directly asking the LLM for a score,
    we ask it to provide a score for each pair of attributes, one from the groundtruth and another from the summary to be evaluated.
We extract 16 attributes derived from a clinical ontology for each summary;
the attributes are shown to define the elements of quality in the context of a clinical discharge summary in a large scale study~\citep{savvopoulos2018development}.

Unlike previous methods such as FActScore~\citep{min2023factscore}, \methodshort does not require a reliable knowledge source to verify each attribute, but instead relies on the pre-specified ontology to capture relevant aspects of a summary. DocLens ~\citep{xie2024doclens},
similar to our work, performs a fine-grained evaluation of medical text using various generation models and shows that such scoring has higher agreement with the judgements of medical experts.
However, DocLens relies on the model to generate the important attributes rather than a clinical ontology. 

% \method is similar to FActScore~\citep{min2023factscore}, which shows that LLM scoring can be improved by decomposing generations into atomic facts and computing the percentage of those facts that can be supported by a knowledge source.
% However, rather than assuming a reliable knowledge source to verify for each atomic fact, we instead extract clinical attributes and score the similarity between each pair.
    
%This structure allows the LLM to be used in a way that best facilitates %finding matches with the groundtruth document.
%This is similar to related works~\citep{}, but the use of a clinical %ontology for attribute selection and the application to clinical %summarization is new here.

We perform experiments on clinical summarization tasks using discharge notes from \mbox{MIMIC-III}~\citep{johnson2016mimic} and various LLMs.
% , including GPT-4~\citep{openai2023gpt4} and LLaMA-3.1~\citep{dubey2024llama}.
We find that that \method{} helps reduce the gap between automated metrics and human annotators.
In addition, \methodshort{} enables auditing the evaluation for a particular summary by providing grounding for each attribute, in the form of a short text span in the input.
% This enables evaluations to be used in a human-in-the-loop setting, helping domain experts to quickly audit the effectiveness of a summary, a potentially necessary step in a high-stakes medical domain.
% we find that the extracted interpretations match human judgements of relevant information, enabling auditing by a human and helping to build a path towards trustworthy evaluation of clinical summarization in resource-constrained scenarios.    

% Additionally, \methodshort{} enables auditing the evaluation for a particular summary, by providing  grounding for each attribute, in the form of a short text span in the input.
% This enables these evaluations to be used in a human-in-the-loop setting, helping domain experts to quickly audit the effectiveness of a summary, a potentially necessary step in a high-stakes medical domain.
% Unlike posthoc feature importance~\citep{lundberg2017unified,ribeiro2016should}, or intrinsically interpretable models~\citep{rudin2021interpretable,singh2023augmenting}, our interpretable grounding we generate comes directly from an LLM, similar to recent works that use LLMs to generate explanations~\citep{rajani2019explain,macneil2022generating,chen2024consistent} and ground those explanations in evidence~\citep{rashkin2021measuring,xue2023rcot,gero2023self}.

%%%%%%%%%%%%%%%%%%%%%%%%%%%%%%%%%%%%%%%%%%%%%%%%%%%%%%%%%%%%%%%%
\section{Methods and experimental setup}

\subsection{Methods: \method}

\cref{fig:intro} shows the various steps of the \method{} (\methodshort) pipeline.
The pipeline takes in two summaries: a groundtruth summary and a generated summary (which we are trying to evaluate).
Additionally, it uses a set of attributes that are important for the task at hand.
These attributes comes from an ontology that utilizes domain knowledge to define the elements of quality, in our case the context of a clinical discharge summary in a large scale study~\citep{savvopoulos2018development}, e.g. \textit{Admission diagnosis} or \textit{Goals of care}.

Each step in the pipeline is achieved by prompting an LLM.
In the first step, the \method{} pipeline first extracts text for each attribute in the ontology from both summaries.
Then, for each pair of extracted attributes, it generates a similarity score by prompting an LLM to compare them on a scale of 1--4 (all prompts are provided in \cref{sec:appendix_prompts}).

In a final (optional) interpretation step, an LLM can be used to ground the scores for each attribute using text spans in the input.
This can be achieved by directly prompting an LLM to find a text span that led to the extracted text for each attribute.
This step is made possible by the \method{} structuring process.

\subsection{Experimental setup}

\paragraph{Dataset}
We analyze discharge summaries for 300 expert selected \mbox{MIMIC-III} notes ~\citep{johnson2016mimic} and release scripts to process and use this data. The notes are selected based on the quality of their discharge summaries, where quality is defined as having most of the important attributes and contain patient data that can be verified from the clinical notes.
To avoid context size bias, we select MIMIC notes with no more than 4,000 tokens (measured with tiktoken's \texttt{p50\_base tokenizer}\footnote{\url{https://github.com/openai/tiktoken}}).

\paragraph{Models}
We evaluate 6 different LLMs:
GPT-4o~\citep{openai2023gpt4} \texttt{gpt4-0314},
GPT-3.5~\citep{ouyang2022training}
\texttt{gpt-3.5-turbo-0613},
all accessed securely through the Azure OpenAI API.
We also evaluate several instruction-finetuned models: LLaMA-2 70B~\citep{touvron2023llama2},
and LLaMA-3.1 [8B, 70B]~\citep{dubey2024llama}.
% We evaluate each LLM both for generating summaries and for scoring summaries.
We compare methods for scoring with two popular metrics:
BERT-Score~\citep{zhang2019bertscore}, and ROUGE-L~\citep{lin2004rouge}.
% We set the sampling temperature for LLM decoding to 0.1.
\begin{figure*}
    \centering
    \includegraphics[width=0.8\textwidth]{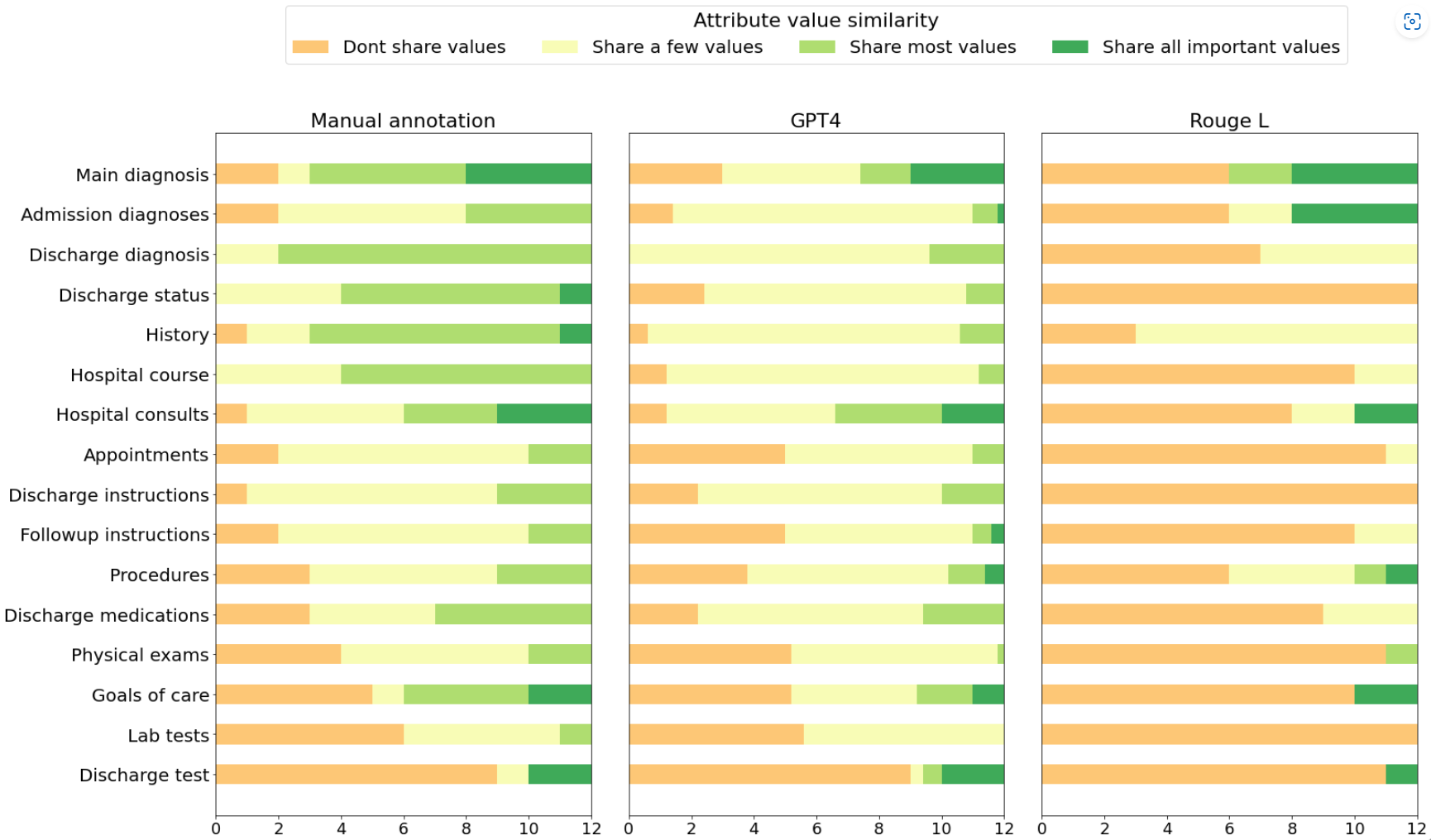}
    \vspace{-5pt}
    \caption{
    Distribution of scores across different attributes.
    \method enables auditing which aspects of discharge summaries are generally accurate (e.g. \textit{Main diagnosis}) and which are often inaccurate (e.g. \textit{Lab test}). 
    }
    \label{fig:sentiment_distr}
\end{figure*}

\paragraph{Annotations and metrics}

On a subset of our selected notes, five blinded human annotators (all are expert medical coders) evaluate the similarities between the extracted text for each attribute.
They score 16 attributes for 24 total documents (12 test summaries generated by GPT-4 and 12 test summaries generated by LLaMA-3 70B),
for a total of 384 attribute labels.
Each label is scored on a 4-point scale: \textit{Not similar}:attributes don't share similar elements, \textit{Somewhat similar}: attributes share a few elements, \textit{Very similar}: attributes share most elements, \textit{Essentially the same}: attributes share all important elements.
We evaluate the match between the manually annotated attribute labels and the automated attribute labels using Pearson correlation, Spearman (rank) correlation, root mean-squared-error (RMSE), and assess inter-annotator agreement with Fleiss' Kappa~\citep{fleiss1971measuring}.

%%%%%%%%%%%%%%%%%%%%%%%%%%%%%%%%%%%%%%%%%%%%%%%%%%%%%%%%%%%%%%%%
\section{Results}

\paragraph{Main result: \methodshort improves evaluation quality}

\begin{figure}[t]
    \centering
    \includegraphics[width=\columnwidth]{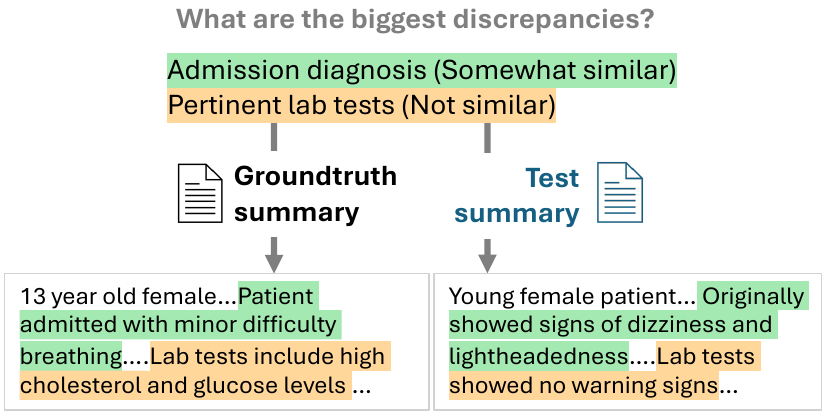}
    \vspace{-15pt}
    \caption{Example interpretation of document similarity.
    Scores for each element of the ontology can be analyzed and grounded in quotes from the source summaries for better understanding.}
    \label{fig:interpretation}
\end{figure}

\cref{tab:main} shows how well different scoring methods match the annotations provided by humans (the inter-annotator Fleiss' Kappa score is 0.71). 
\method yields a considerable improvement for all metrics.
Scoring with GPT-4 yields the best match with human annotators both in Pearson and Spearman correlations and has the least RMSE error. Even a traditional scoring metric, ROUGE-L, shows significant improvement when attributes are structured before scoring.
% and \method yields a sizable improvement.  

% , both markedly surpassing baseline scoring methods in alignment with human annotations.
The enhanced correlations across all metrics substantiate that attribute-based, fine-grained scoring outperforms holistic methodologies.

% \looseness=-1 
\cref{fig:sentiment_distr} shows the breakdown of similarities across attributes for different metrics.
Some attributes, such as \textit{Main diagnosis} are more often correctly identified in the test summaries, whereas others, such as \textit{Lab test} and \textit{Discharge test} are rarely correct.
Aggregating these attribute evaluations through \methodshort allows for improved evaluation compared to evaluating an entire summary at once.

\begin{table}[t]
    \centering
    \small
    \caption{Main result: \method improves the match between automated scoring metrics and human annotators.
    For all methods, attribute structuring is performed by prompting GPT-4.
    }
    
% \begin{tabular}{lcc|ccc}
% \toprule
% & MAE ↓                & RMSE ↓      & Pearson ↑   & Spearman ↑  \\
% \midrule
% & Default/\textbf{\methodshort} & Default/\textbf{\methodshort} & Default/\textbf{\methodshort} & Default/\textbf{\methodshort}\\
% \midrule
% Rouge-1 & \textbf{0.15 / 0.15} &\textbf{ 0.17 / 0.17} & 0.36 / \textbf{0.64} & 0.34 / \textbf{0.48} \\
% Rouge-L & 0.24 / \textbf{0.17}          & 0.26 / \textbf{0.18} & 0.35 / \textbf{0.59} & 0.32 / \textbf{0.44} \\
% BERT-Score & 0.47 / \textbf{0.16}          & 0.48 / \textbf{0.18} & 0.17 / \textbf{0.71} & 0.23 / \textbf{0.69} \\
% GPT-Score & 0.15 /\textbf{ 0.06}          & 0.18 / \textbf{0.07} & 0.46 / \textbf{0.84} & 0.38 / \textbf{0.77}\\
% \bottomrule
% \end{tabular}
\begin{tabular}{lc|cc}
\toprule
& RMSE ↓      & \makecell{Pearson\\correlation} ↑   & \makecell{Spearman\\correlation} ↑  \\
\midrule
& Default/\textbf{\methodshort} & Default/\textbf{\methodshort} & Default/\textbf{\methodshort}\\
\midrule
ROUGE-L &  0.17 / \textbf{0.12} & 0.48 / \textbf{0.70} & 0.36 / \textbf{0.70} \\
BERT-Score & 0.52 / \textbf{0.22} & 0.13 / \textbf{0.70} & 0.28 / \textbf{0.80} \\
GPT-4 & 1.10 / \textbf{0.06} & 0.54 / \textbf{0.91} & 0.43 / \textbf{0.81}\\
\bottomrule
\end{tabular}
    \label{tab:main}
\end{table}

\begin{table}[t]
    % \footnotesize
    % \small
    \centering
    \caption{
    Automatic evaluation scores using different LLMs for generating summaries.
    % and for evaluating.
    % The \textit{Benchmark} column uses Mixtral for structuring summaries as well as evaluating them, making it fully reproducible using open source tools.
    % The remaining columns use GPT-4 for structuring and different LLMs for scoring the attributes.
    % The relative rankings of different LLMs generators is preserved across different columns.
    }
    \begin{tabular}{lll}
\toprule
Summarizer     & \methodshort Score  \\
\midrule
LLaMA-2 (70B) & 82.8   \\
% Phi-3-med   & 57.6   \\
GPT-3.5     & 90.0   \\
LLaMA-3.1 (8B) & 91.8   \\
GPT-4-0314  & 93.4   \\
GPT-4o     & 95.3  \\
LLaMA-3.1 (70B) & 95.7   \\
\bottomrule
\end{tabular}
    \label{tab:vary_generator}
\end{table}

\paragraph{Analysis: \methodshort scores can be used to evaluate the quality of summaries}In settings where there aren't manual annotations, we can use \methodshort score to evaluate the quality of summaries generated by LLMs. In \cref{tab:vary_generator}, we can see that \methodshort scores for summaries generated by six LLMs are consistent with the expected quality of the individual models.
% We now evaluate how well different LLMs perform under an automatic evaluation using \method.
% The first column of \cref{tab:vary_generator} shows the scores achieved by different models;
% the scores rely only on the open-source Mixtral model to perform both structuring and scoring, enabling full reproducibility.

% GPT-4 achieves the highest score, followed by GPT-3.5 and Mixtral (8x7B).
% We additionally evaluate how consistent the scores are with scores provided using different language models for structuring and scoring.
% The right section of \cref{tab:vary_generator} shows evaluation scores when structuring with GPT-4 and scoring with various different LLMs.
% We find that the absolute scores in different columns vary widely, but the relative ordering of LLM generators is almost perfectly preserved between the benchmark and all other settings (the only exception is that the Mixtral scorer slightly prefers the Mixtral generator to the GPT-3.5 generator).
% Regardless of the LLM used for scoring,
% Different scoring models yield different ranges of scores,
% so comparisons should always be made using a fixed scorer.

\paragraph{Analysis: \method facilitates interpretation}
Going beyond aggregate interpretations,
\cref{fig:interpretation} shows an example of how similarity can be audited for a single pair of summaries.
Unlike posthoc feature importance~\citep{lundberg2017unified,ribeiro2016should}, or intrinsically interpretable models~\citep{rudin2021interpretable,singh2023augmenting}, our interpretable grounding comes directly from an LLM, similar to recent works that use LLMs to generate explanations~\citep{macneil2022generating,chen2024consistent} and ground those explanations in evidence~\citep{xue2023rcot,gero2023self}.
Since \methodshort decomposes similarity into a series of attributes, the attributes can each be audited by directly prompting an LLM (here, GPT-4) to find a text span that serves as evidence for the extracted attribute in each document.
\cref{fig:interpretation} shows text spans for two discrepancies: \textit{Admission diagnosis} and \textit{Pertinent lab tests}, which are only partially
retained in the test summary.

\section{Discussion}

\method{} constitutes an important step towards safely deploying LLMs in healthcare settings.
As LLMs continue to generally improve,
their ability to produce and evaluate clinical summaries with \methodshort is likely to improve as well.
Going forward, \method can be harnessed in a variety of ways to improve clinical NLP summarization, 
% what is studied here, \eg for studying clinical decision rules~\citep{kornblith2022predictability},
e.g. for evaluating free-form text such as conversations~\citep{tu2024towards} or
% e.g. for clinical decision support systems~\citep{liu2023assessing},
improving model distillation by helping to filter effective summaries~\citep{wu2023pmc,toma2023clinical}.
% and improving the explainability of the process~\citep{singh2024rethinking}.

% \paragraph{Limitations}
One limitation of \method is that it incurs a high computational cost as multiple LLM calls are chained together;
however, these costs may continue to decrease as models become more efficient~\citep{dao2022flashattention}.
Another limitation is that LLMs (and thereby \method) continue to be sensitive to prompts,
increasing the need for methods to make LLMs more amenable to prompting~\citep{ouyang2022training,zhou2023context,zhang2023tell} and to make finding strong prompts easier~\citep{shin2020autoprompt,singh2023explaining,xu2023k,morris2023tree}.
Additionally, \methodshort relies on a human-given ontology for evaluation, but could be made more general by developing methods to automatically define extract and define an ontology for a given domain.
% While \method helps to improve the automatic evaluation of summaries, there are nevertheless risks present with fully relying on LLMs without human supervision.

{
    \small
    \bibliography{refs}
    % \bibliography{jmlr-sample}
}

\clearpage
\appendix
\FloatBarrier
\renewcommand{\thefigure}{A\arabic{figure}}
\renewcommand{\thetable}{A\arabic{table}}
\setcounter{table}{0}
\setcounter{figure}{0}

\section{Appendix}
\label{sec:appendix_prompts}

\begin{table*}[ht]
    \footnotesize
    \small
    \centering
    \caption{
    Automatic evaluation scores using different LLMs for generating and evaluating summaries.
    % The \textit{Benchmark} column uses Mixtral for structuring summaries as well as evaluating them, making it fully reproducible using open source tools.
    All use GPT-4 for structuring and different LLMs for scoring the attributes.
    The relative rankings of different LLMs generators is preserved across different columns.
    Results are for a subset of 30 randomly chosen notes.
    }
    \begin{tabular}{llc|ccccclc}

\toprule
\vspace{3pt}
& & \multicolumn{7}{c}{Scorer (with GPT-4 structuring)} \\
& & \textbf{AVG}& \makecell[t]{LLaMA-2\\(7B)} & \makecell[t]{Mistral\\(7B)} & \makecell[t]{LLaMA-2\\(70B)} & \makecell[t]{Mixtral\\(8x7B)} & GPT-3.5 & GPT-4 \\
\midrule
\multirow{6}{*}{\rotatebox[origin=c]{90}{Generator}} & LLaMA-2 (7B)  &45.97 & 65.4         & 47.1    & 49.0          & 38.5          & 40.3    & 35.5 \\
& Mistral    &46.37    & 66.1         & 47.7    & 49.2          & 38.7          & 40.8    & 35.7 
 \\
& LLaMA-2 (70B) &49.70  & 66.7         & 50.6    & 54.3          & 43.4          & 43.6    & 39.6 
 \\
& Mixtral (8x7B)&51.30  & 68.4         & 52.4    & 54.9          & 46.0          & 44.9    & 41.2 
 \\
& GPT-3.5     &52.78    & 73.1         & 53.7    & 56.7          & 45.9          & 45.7    & 41.6 
 \\
& GPT-4     & \textbf{56.02}    & \textbf{73.8}         & \textbf{56.9}    & \textbf{61.2}          & \textbf{51.2}          & \textbf{48.8}    & \textbf{44.2}
 \\
\bottomrule
\end{tabular}

    \label{tab:vary_generator_summarizer}
\end{table*}

\begin{table*}[ht]
    \footnotesize
    \small
    \centering
    \caption{
    Automatic evaluation scores using different LLMs for structuring summaries yield similar rankings between generator models, although GPT-4 structuring generally yields higher scores.
    Both use Mixtral (8x7B) for scoring.
    Results are for a subset of 30 randomly chosen notes.
    }
    \begin{tabular}{lcc}
\toprule
Generator & Mixtral structuring & GPT-4 structuring \\
 \midrule
LLaMA-2 (7B)   & 36.2              & 38.5              \\
Mistral        & 37.4              & 38.7              \\
LLaMA-2 (70B)  & 38.4              & 43.4              \\
Mixtral (8x7B) & 39.2              & 46.0              \\
GPT-3.5        & 39.4              & 45.9              \\
GPT-4          & \textbf{45.1}     & \textbf{51.2}     \\
 \bottomrule
\end{tabular}
    \label{tab:vary_structuring}
\end{table*}

\subsection{Prompts}

Here we include the prompts we used to:
\begin{itemize}
 \item Generate summaries
 \item Structure to extract attributes ,and 
 \item Score each pair of attributes.
 \item A prompt to score without structuring as a baseline. 
\end{itemize}
\textbf{1. Generation prompt}: \\
Prompt = \textit{``You are tasked with generating a high quality clinical discharge summary for the provided input text INPUT\_DOC. The summary has to be very relevant to the input document and cover the most important aspects of the input. Follow the following steps:} \\
\textit{
  Step 1: Generate the most common sections that usually appear in a clinical discharge summary.}
  \textit{Step 2: Not all the sections from Step 1 will have content. See which of those sections have contents from the INPUT\_DOC. Remove the sections that don't have information.} \\
  \textit{Step 3 : Use the following concepts to generate appropriate content. These are not Sections. Concepts:
    patient information and service type, diagnosis given at the time of admission,  brief summary of initial presentation and diagnostic evaluation, pertinent physical findings relevant to diagnoses, goals of care; level of treatment,code status(e.g. curative,life-prolonging palliative, and symptomatic palliative), course in hospital; synotpic,problem-based description of sequential events and respective evaluations, treatments, and prognoses, hospital consults; description of specialty and/or allied health consults, procedures in hospital; a list of procedures with key findings and date, principal discharge diagnosis or main reason for admission and all additional pertinent diagnoses where applicable,  discharge medications with specific description of new, altered, and discontinued medications and rationale for changes,  lab tests and investigative results, tests ordered during the hospitalization that are pending at the time of discharge, outcome of care/condition at discharge; sense of the patient health status at discharge includes functional status, and cognitive status, outstanding issues for follow-up and recommendations to a recipient health-care provider during discharge, appointments after discharge including person responsible for scheduling, care provider, discharge instructions; list of information/education provided to the patient during discharge, main author of the discharge summary or attending clinician. } \\

   \textit{Step 4: Put the generated content in a coherent order. Format in such a way that the dishcarge summary has an excellent coherence, fluency, and consistency. Remember to use the sections you generated in Step 1, don't make the concepts in Step 2 as a section, they are just suggestive concepts not sections. } \\

   \textit{Step 5: Remove sections with no information. Dont put 'not specified' or 'not mentioned' or 'none specified' in a section. Just remove everything for that section including the section header.} \\ 
   \textit{Step 5: Return the final discharge summary with all the remaining sections that have contents. Remember to remove sections with no information.
Context : "}

\textbf{2. Structuring prompt}: \\
We use ResponseSchema from langchain to inform the model what each attribute description is. 
\begin{description}
\item[ad\_diag\_schema] = ResponseSchema(name = 'ad\_diag', description='preliminary or working diagnosis given at the time of admission')
\item[dc\_diag\_schema] = ResponseSchema(name = 'dc\_diag', description='the list of principal discharge diagnosis or main reason for admission and all additional pertinent diagnoses where applicable')
\item[main\_diag\_schema] = ResponseSchema(name = 'main\_diag', description='diagnosis mostly accountable for the largest portion of the patient"s stay, responsible for the greatest part of the length of stay ')
\item[history\_schema] = ResponseSchema(name = 'history', description='a brief summary of initial presentation and diagnostic evaluation')
\item[physical\_schema] = ResponseSchema(name = 'physical', description='pertinent physical findings relevant to diagnoses ')
\item[goals\_schema] =ResponseSchema(name = 'goals', description='goals of care; level of treatment,code status(e.g. curative,life-prolonging palliative, and symptomatic palliative)') 
\item[course\_schema] = ResponseSchema(name = 'course', description='course in hospital; synotpic,problem-based description of sequential events and respective evaluations, treatments, and prognoses ')
\item[consults\_schema] = ResponseSchema(name = 'consults', description='hospital consults; description of specialty and/or allied health consults')
\item[procedures\_schema] = ResponseSchema(name = 'procedures', description='procedures in hospital; a list of procedures with key findings and date')
\item[ds\_med\_schema] = ResponseSchema(name = 'ds\_med', description='a list of all discharge medications with specific description of new, altered, and discontinued medications and rationale for changes')
\item[lab\_schema] = ResponseSchema(name = 'lab', description='pertinent lab tests and investigative results')
\item[ds\_test\_schema] = ResponseSchema(name = 'ds\_test', description='tests ordered during the hospitalization that are pending at the time of discharge ')
\item[ds\_status\_schema] = ResponseSchema(name = 'ds\_status', description='outcome of care/condition at discharge; sense of the patient health status at discharge includes functional status, and cognitive status')
\item[followup\_schema] = ResponseSchema(name = 'followup', description='outstanding issues for follow-up and recommendations to a recipient health-care provider during discharge')
\item[appt\_schema] = ResponseSchema(name = 'appt', description='appointments after discharge including person responsible for scheduling, care provider ')
\item[instruct\_schema] = ResponseSchema(name = 'instruct', description='discharge instructions; list of information/education provided to the patient during discharge')
\item[author\_schema] = ResponseSchema(name = 'author', description='main author of the discharge summary or attending clinician')
\end{description}

\textit{
prompt = ``You are an expert in information extraction and structuring from clinical notes.  Given a clinical note, create a structured output. For a given variable, if you can not determine/find a value, return NONE. Dont add any extra text, just the structured value. Here is the note: {INPUT\_DOC}. {format\_instructions}
"} 

\textbf{3. Structured attribute Scoring Prompt:} \\

prompt = ``You will be given a python dictionary containing a clinical variable as a key and a list containing two values for the variable as values.

Your task is to rate how similar the values are given the variable. Compare value1 and value2 for semantic similarity(similarity in meaning) given the variable and the criteria below. Two values can be very similar in meaning even if they are phrased differently. Also remember that this is a clinical document, take that into account.
When scoring the similarity between two clinical terminologies, assign a value from 1 to 4, where 1 signifies a lack of similarity and 4 indicates identical meanings. Consider the context, clinical relevance, and semantic alignment between the terminologies. If the terminologies convey vastly different meanings, assign a score of 1. A score of 2 is appropriate for terminologies that, while related, represent different concepts or emphasize distinct elements. A score of 3 should be used for terminologies with substantial semantic overlaps but minor differences. Finally, if the terminologies are semantically equivalent and interchangeable, warranting no clinical distinction, assign a score of 4. Ensure that the assessment reflects the degrees of similarity in meaning, irrespective of syntactical differences, context, or minor variances in expression.

Please make sure you read and understand these instructions carefully.
Return the similarity score. Return only the score, don't include any other extra text."

\textbf{4. Document scoring without structuring prompt}

prompt = ``You will be given two clinical discharge summaries, summary1 and summary2.

Your task is to rate how similar the two summaries are from 1-10. 1 is not similar while 10 is essentially identical. Focus on the following important clinical variables when performing the comparison:
  How similar is the diagnosis, how similar are the goals, how similar is hospital course and history, how similar are the medications administered, how similar are the physical condition diagnoses, how similar are the followup consults and procedures, how similar are the lab tests performed, how similar are the patients discharge status, is the follow-up instructions similar, are there any similar appointments, and instructions
only return the score, dont add any extra text
Here are the inputs discharge summaries:
 summary1: [value1]
 summary2: [value2]."

% \FloatBarrier
\subsection{Prompts given to human annotators}

\begin{figure}
    \centering
    \includegraphics[width=\columnwidth]{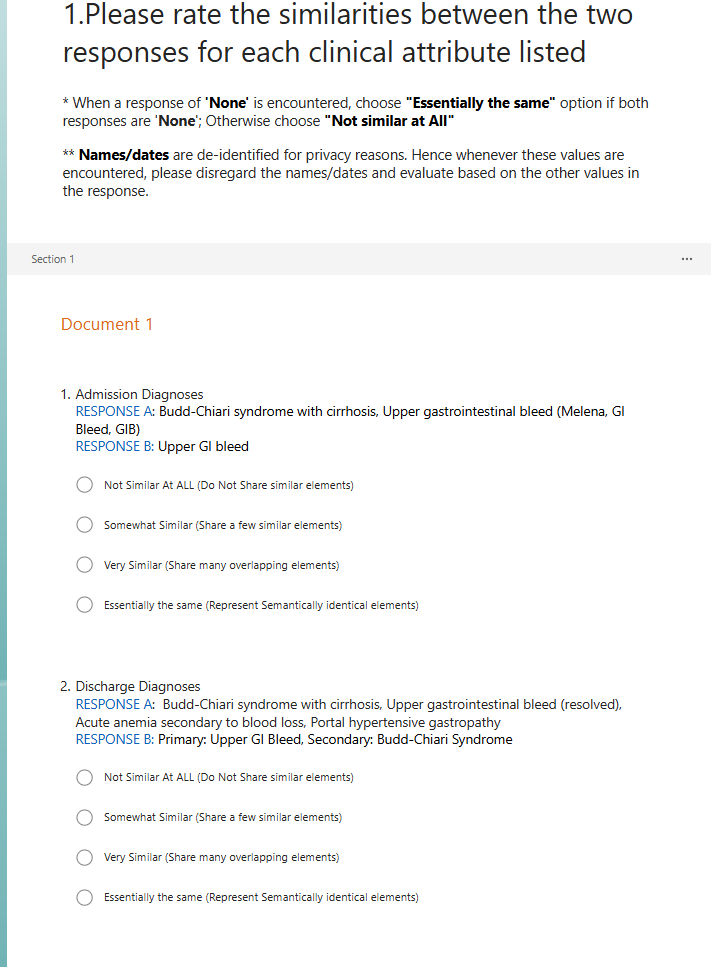}
    \caption{Annotation interface screenshot.}
    \label{fig:annotation}
\end{figure}

\cref{fig:annotation} shows a sample interface given to human annotators.
Human annotators were presented values for attributes through a web interface along with the prompt ``For each attribute listed on the left, two possible values are provided(different colors). Please rate the semantic similarity between the values.''.
Annotators are three machine learning researchers that are authors of the current study.

% \appendix

% \section{First Appendix}\label{apd:first}

% This is the first appendix.

% \section{Second Appendix}\label{apd:second}

% This is the second appendix.

\end{document}